\pdfoutput=1

\documentclass[11pt]{article}

\usepackage[]{ACL2023}

\usepackage{times}
\usepackage{latexsym}

\usepackage[T1]{fontenc}

\usepackage[utf8]{inputenc}

\usepackage{microtype}

\usepackage{inconsolata}

\usepackage{enumitem}
\setitemize{noitemsep, topsep=0pt, parsep=0pt, partopsep=0pt}

\usepackage{dsfont}
\usepackage{amsmath, amssymb}
\usepackage{graphicx}

\usepackage{booktabs}

\usepackage{arydshln}
\title{Language Anisotropic Cross-Lingual Model Editing}

\author{Yang Xu \qquad Yutai Hou \qquad Wanxiang Che \qquad Min Zhang \\
	Harbin Institute of Technology \\
	\texttt{\{yxu, ythou, car\}@ir.hit.edu.cn \qquad zhangmin2021@hit.edu.cn}\\}

\begin{document}
	\maketitle
	\begin{abstract}
		Multilingual pre-trained language models can learn task-specific abilities or memorize facts across multiple languages but inevitably make undesired predictions with specific inputs.
		Under similar observation, model editing aims to post-hoc calibrate a model targeted to specific inputs with keeping the model's raw behavior.
		However, existing work only studies the monolingual scenario, which lacks the \textit{cross-lingual transferability} to perform editing simultaneously across languages.
		In this work, we focus on cross-lingual model editing.
		Firstly, we define the cross-lingual model editing task and corresponding metrics, where an edit in one language propagates to the others.
		Next, we propose a framework to naturally adapt monolingual model editing approaches to the cross-lingual scenario using parallel corpus.
		Further, we propose \textit{language anisotropic} editing to improve cross-lingual editing by amplifying different subsets of parameters for each language.
		On the newly defined cross-lingual model editing task, we empirically demonstrate the failure of monolingual baselines in propagating the edit to multiple languages and the effectiveness of the proposed \textit{language anisotropic} model editing.
		Our code is publicly available at \url{https://github.com/franklear/LiME}.
	\end{abstract}
	
	\section{Introduction}
	Pre-trained language model based approaches have become the best practice in many fields, including multilingual NLP~\citep{ptm-book, tunstall2022natural}.
	During training, Transformer-based~\citep{DBLP:conf/nips/VaswaniSPUJGKP17} models can embed language abilities~\citep{DBLP:conf/emnlp/GevaSBL21} and memorize facts~\citep{dai-etal-2022-knowledge} in the parameters.
	Though, models inevitably make undesired predictions with specific inputs, such as mistake labels or outdated facts.
	Moreover, the performance of multilingual models is unbalanced across languages, leading to inconsistency predictions over the same input in different languages.
	However, the high cost of training and data collecting makes it unrealistic to re-train the models using calibrated data in all languages.
	Therefore, there is a pressing need for an approach to calibrate multilingual pre-trained models across all languages of interest simultaneously.

	\begin{figure}[t]
		\centering
		\includegraphics[width=\columnwidth]{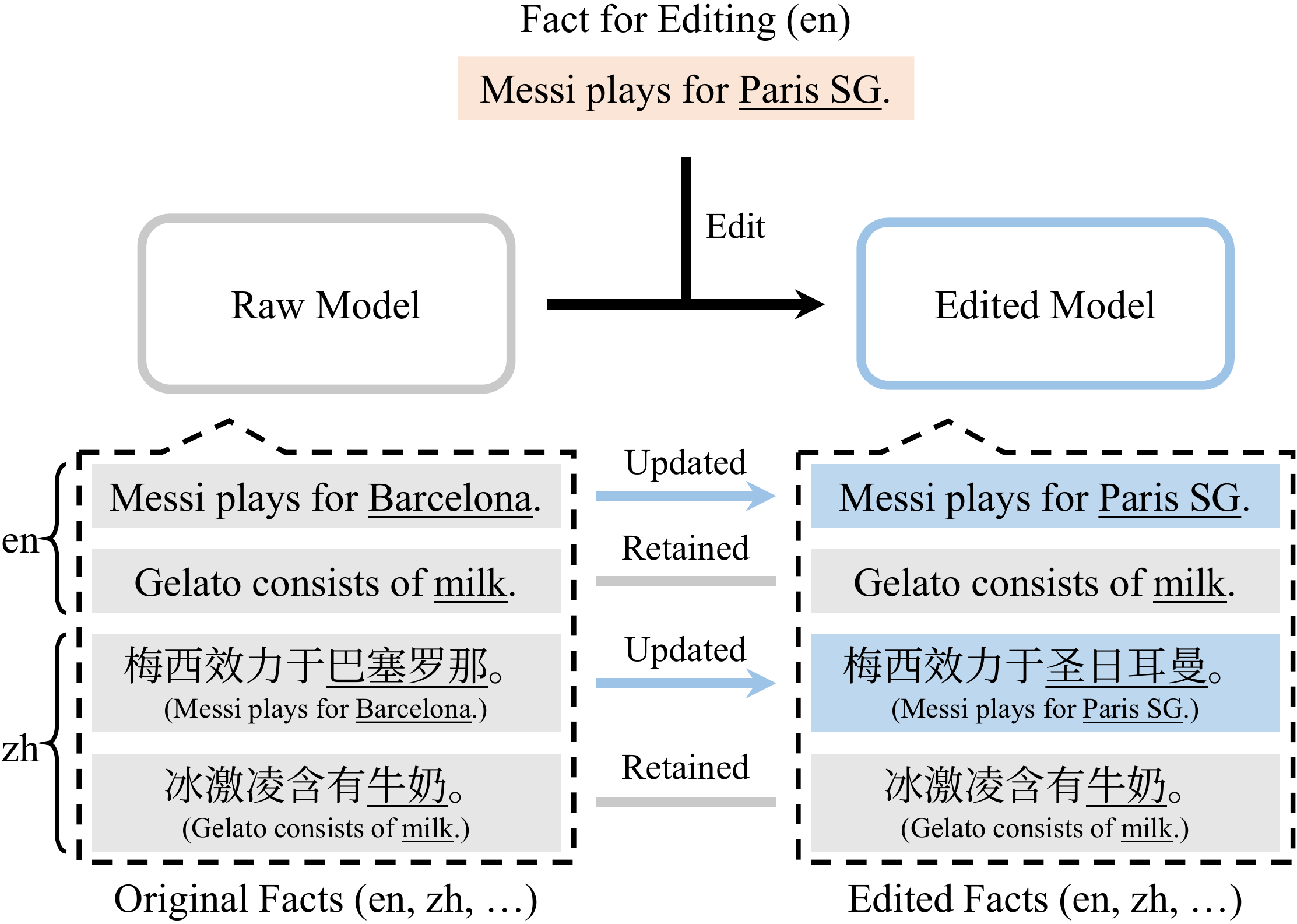}
		\caption{
			An example of cross-lingual model editing for updating facts, where we represent facts inferred from models as sentences in the dashed boxes.
			The goal is to update the given fact while retaining unrelated facts. Further, cross-lingual editing requires the edit in one language (e.g., en) to affect all languages (en, zh, \dots).
		}
		\label{fig:task}
	\end{figure}
	
	As an emerging research area, model editing allows us to calibrate the behavior of pre-trained language models targeted to specific inputs~\citep{DBLP:conf/iclr/SinitsinPPPB20, DBLP:conf/emnlp/CaoAT21, DBLP:conf/iclr/MitchellLBFM22, DBLP:conf/icml/MitchellLBMF22, DBLP:journals/corr/abs-2202-05262, DBLP:journals/corr/abs-2210-07229, DBLP:journals/corr/abs-2111-13654}.
	However, challenges emerge when applying model editing to the cross-lingual scenario, due to the two features of multilingual pre-trained models:

	The first is \textit{cross-lingual transferability}. Based on prior research conducted on pre-trained multilingual models like XLM~\citep{DBLP:conf/nips/ConneauL19} and InfoXLM~\citep{DBLP:conf/naacl/ChiDWYSWSMHZ21}, it is well-established that incorporating diverse language data during training leads to advantageous cross-lingual transfer effects.
	Thus, input with the same meaning can be expressed in multiple languages as completely different sentences.
	The editor has to be aware of this feature in case it suffers from editing failure in unseen languages.
	
	The second is \textit{language anisotropy}. Recent work reveals that language-specific and language-universal parameters exist in the multilingual pre-trained model~\citep{DBLP:conf/emnlp/WangLT20}.
	This finding means the model tends to mainly activate a subset of its parameters depending on the language to be processed, which we call \textit{language anisotropy}.
	An editor which treats all parameters identically for all languages is not \textit{language anisotropic}, potentially harming other languages when editing.

	In this work, we propose for the first time cross-lingual model editing on multilingual pre-trained language models.
	Different from existing model editing, an edit in a single language propagates to the others in cross-lingual model editing.
	As is shown in Figure~\ref{fig:task}, with cross-lingual model editing, editing a fact in English also affects the Chinese version, while retaining unrelated facts.
	
	We propose a simple yet effective framework to adapt existing monolingual model editing approaches to the cross-lingual scenario using the parallel corpus.
	Specifically, we replace the inputs for editor training with their parallel expressions in random languages.
	For example, the editor can be asked to edit model predictions on English input. The edited model is then supervised to enforce that the predictions are updated on parallel Chinese input and retained on unrelated French inputs. The next time, the above languages randomly change.
	To this end, the cross-lingual training formula helps the editor gain \textit{cross-lingual transferability}.
	
	Besides, we leverage the \textit{language anisotropy} nature of the multilingual models to further improve cross-lingual model editing.
	Specifically, we propose to add a group of $L_0$ constrained language-specific masks as the editor's parameters.
	During editing, the masks are used to instruct the editor to focus on different parts of the raw model's parameters according to the inputs' language.
	Training along with the masks, the editor gains the skill of making \textit{language anisotropic} edits.

	Our primary contributions are as follows:
	\begin{itemize}
		\item We define the cross-lingual model editing task and corresponding evaluation metrics.
		\item We propose a simple yet effective framework to adapt the monolingual editing approaches to the cross-lingual scenario.
		\item We propose \textit{language anisotropic} model editing to improve cross-lingual model editing significantly.
	\end{itemize}

	\section{Background: Model Editing}

	\citet{DBLP:conf/iclr/SinitsinPPPB20} propose Editable Training (model editing) as an efficient approach to modify the behavior of a trained model on specific inputs, where three core requirements are highlighted:
	\begin{itemize}
		\item \textit{Reliability}: the edited model acquires the desired behavior on specific inputs.
		\item \textit{Locality}: the edit influences the model on other inputs of interests as little as possible.
		\item \textit{Efficiency}: the editing approach should be computationally efficient.
	\end{itemize}

	\textit{Reliability} and \textit{locality} are essential attributes of the model editing task, while \textit{efficiency} is required to make the editor usable.

	Recent work explores several ways to solve the model editing problem~\citep{DBLP:conf/iclr/SinitsinPPPB20, DBLP:conf/iclr/MitchellLBFM22, DBLP:journals/corr/abs-2202-05262, DBLP:journals/corr/abs-2210-07229}.
	Despite the variety of algorithms, their training formulas are similar, i.e., training the editor end-to-end on editing data under the condition of \textit{reliability} and \textit{locality}.
	Specifically, a training step of the editor contains two stages.
	1) Editing stage: the editor is used to edit desired predictions into the raw model $f(\cdot ; \boldsymbol{\theta})$, producing the edited model $f(\cdot ; \boldsymbol{\theta}_u)$.
	2) Editor training stage: the edited model is then constrained under the requirements of \textit{reliability} and \textit{locality}, corresponding to two core objectives respectively.
	
	For \textit{reliability}, the edited model need to make the desired prediction $y_e$ in response to the input $x_e$.
	This requirement refers to the task loss $L_{\textrm{task}}$, e.g., cross-entropy or $L_2$.
	So we have
	\begin{equation}
		\label{eq:loss rel}
		\tag{$L_{\textrm{rel}}$}
		L_{\textrm{rel}} = \lambda_{\textrm{rel}} L_{\textrm{task}} \left( f(x_e; \boldsymbol{\theta}_u), y_e \right) \text{.}
	\end{equation}
	
	For \textit{locality}, the edited model needs to retain predictions of unrelated inputs, which means that for an unrelated input $x_r$, the output $f(x_r; \boldsymbol{\theta})$ should be kept.
	Though a similar loss like $L_{\textrm{rel}}$ can work in theory, the stronger KL divergence loss is used to minimize the side effect on unrelated labels
	\begin{equation}
		\label{eq:loss loc}
		\tag{$L_{\textrm{loc}}$}
		L_{\textrm{loc}} = \lambda_{\textrm{loc}} \operatorname{KL} \left( f(x_r; \boldsymbol{\theta}_u) \parallel f(x_r; \boldsymbol{\theta}) \right) \text{.}
	\end{equation}
	
	In addition, other auxiliary objectives can be utilized which do not affect the training formula.
	
	Note that the goal is to train the editor instead of the raw model.
	During training, the gradients propagate through the edited model to the editor.
	At test time, only the editing stage is needed.
	Overall, the training of the editor is a meta version of the model training because the ``data'' that the editor processes is the model (plus the input-prediction pair to be edited).
	
	\section{Cross-Lingual Model Editing}

	\subsection{Task Definition}
	\label{sec:Cross-Lingual Model Editing:Task Definition}
	
	Following the work on monolingual model editing, we continue taking the idea of making an edit with \textit{reliability} and \textit{locality}~\citep{DBLP:conf/iclr/SinitsinPPPB20}, while introducing \textit{cross-lingual transferability}.
	
	\begin{figure*}[h]
		\centering
		\includegraphics[width=\linewidth]{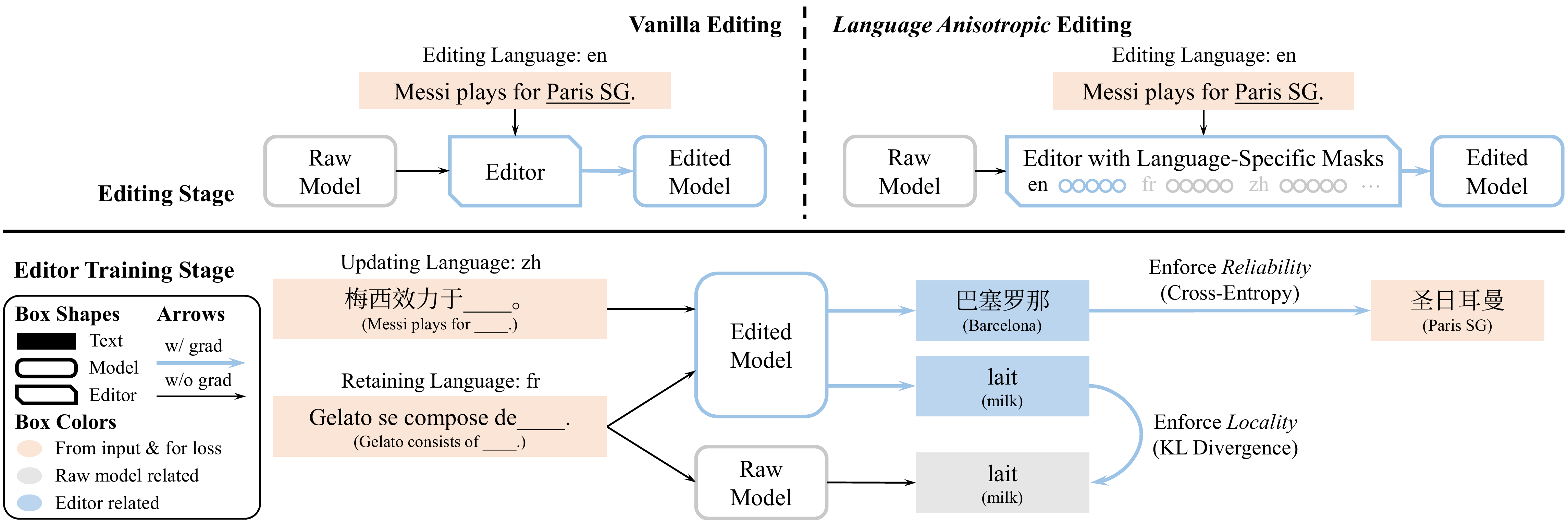}
		\caption{
			The overall framework of the proposed cross-lingual model editing.
			Each training step consists of two stages. The editor edits the model at first, then losses for \textit{reliability} and \textit{locality} are obtained from the outputs of the edited model to supervise the editor.
			Languages of editing/updating/retaining are randomly sampled in each training step to endow the editor with \textit{language transferability}.
			Our novel \textit{language anisotropic} model editing applies soft masks according to the editing language, which are supervised using the re-parameterized $L_0$ loss.
		}
		\label{fig:main}
	\end{figure*}
	
	Assuming we have a model $f$ parameterized by $\boldsymbol{\theta}$ that maps the input $x$ to the prediction $p = f(x; \boldsymbol{\theta})$. An update is needed when we want the model change its prediction from $p$ to $y$.
	Here the requirement of \textit{cross-lingual transferability} brings the key difference.
	The same input can be represented in multiple languages, producing parallel sentences.
	Therefore, the edit with \textit{reliability} for $x$ should affect the parallel inputs, denoted as $I(x)$.
	As the example in Figure~\ref{fig:task}, ``Messi plays for Paris SG.'' in English is parallel to its Chinese translation.
	For \textit{locality}, the side effect should be as low as possible, which means the prediction of input $x' \notin I(x)$ is retained.

	Note that under this setting, one edit is always independent of another. The editor revisits the $\boldsymbol{\theta}$ for every edit, then produces the corresponding $\boldsymbol{\theta}_u$.
	Formally, the goal of the editor is to
	\begin{align*}
		\textit{find} \quad & \boldsymbol{\theta}_u,  \\
		\textit{s.t.} \quad & \left\{
		\begin{aligned}
			f(x_p; \boldsymbol{\theta}_u) & = y & \forall x_p \in I(x)  \\
			f(x_n; \boldsymbol{\theta}_u) & = f(x_n; \boldsymbol{\theta}) & \forall x_n \notin I(x)
		\end{aligned}
		\right. , \kern-\nulldelimiterspace & \hphantom{-1} \\
		\textit{given} \quad & x, I(x) = \{x' \vert x'\ \textrm{is parallel to}\ x\}, y, f, \boldsymbol{\theta}.
	\end{align*}
	
	\subsection{Cross-Lingual Editing Based on Monolingual Approaches}
	\label{sec:Cross-Lingual Model Editing:Cross-Lingual Editing Based on Monolingual Approaches}

	Dispite the \textit{cross-lingual transferability}, the requirements of \textit{reliability} and \textit{locality} stay the same with monolingual model editing, which are defined by the training data.
	To fully leverage the monolingual editing approaches and build reasonable baselines, we propose a framework to adapt them to the cross-lingual scenario using the parallel corpus as illustrated in Figure~\ref{fig:main}.
	
	What we need is a slight change in the training formula of the monolingual editing approaches, namely aligning inputs in different languages.
	Given $x_e$ in the editing language $l_e$ as the input to be edited and the corresponding desired prediction $y_e$, the inputs used in the training objectives are sampled over the parallel inputs set $I(x_e)$.
	
	For \textit{reliability}, the edited model is asked to update the prediction to $y_e$ on the sampled input $x_u \in I(x_e)$ in the updating language $l_e$.
	Thus the \textit{reliability} loss (\ref{eq:loss rel}) is modified by replacing $x_e$ with $x_u$.
	For \textit{locality}, the sampled input $x_r \notin I(x_e)$ in the retaining language $l_r$ is used as input, and the \textit{locality} loss (\ref{eq:loss loc}) remains the same.
	
	Monolingual editing is a degenerate case where only a single language is considered, i.e., $l_e = l_u = l_r$. When the languages differ, the editor trained under the above sampling strategy acquires \textit{cross-lingual transferability}.
	
	Intuitively, the editor functions as updating on identical inputs while not affecting unrelated inputs.
	In the above cross-lingual adaptation, \textit{reliability} loss tells the editor what should be identical, and \textit{locality} loss tells what should be unrelated. Thus, the two losses illustrate a semantically equivalent range for the editor across multiple languages, deriving the \textit{cross-lingual transferability}.
	Therefore, the adaptation we make leverages the parallel corpus to inspire the potential of transferability that comes with the model editing task.
	
	\subsection{Language Anisotropic Editing}
	\label{sec:Cross-Lingual Model Editing:Language Anisotropic Editing}
	
	A multilingual pre-trained model like mBERT~\citep{DBLP:conf/naacl/DevlinCLT19} can integrate over one hundred languages in a single model.
	However, a known phenomenon called the curse of multilinguality~\citep{DBLP:conf/acl/ConneauKGCWGGOZ20} exposes the trade-off between the number of languages and the model capacity, implying the languages tend to compete for the shared parameters.
	Further, it is revealed that language-specific and language-universal parameters exist in the multilingual model, which potentially harm its \textit{cross-lingual transferabillity}~\citep{DBLP:conf/emnlp/WangLT20}.
	All this evidence indicates that the multilingual model is \textit{language anisotropic} in the perspective of the parameters.
	Therefore, we introduce a priori, i.e., the update should focus more on a certain subset of parameters according to the language of the input to edit.
	Nevertheless, identifying which language prefers which parameters is not so direct. Our idea is to drive the editor to find the important parameters during training.
	
	As shown in the top-right part of Figure~\ref{fig:main}, we realize the idea with a group of learnable language-specific masks.
	The model editor produces new parameters to update the raw model, so we mask the input/output of the editor to apply an adaptive weighting.
	For an update in language $l$, we mask each parameter (tensor) $\boldsymbol{W}$ to be updated with $\boldsymbol{m}^l_{\boldsymbol{W}} \in \left[ 0, 1 \right]^{\dim \boldsymbol{W}}$ through
	\[\operatorname{mask} (\boldsymbol{W}, \boldsymbol{m}^l_{\boldsymbol{W}}) = \boldsymbol{W} + \boldsymbol{m}^l_{\boldsymbol{W}} \odot \boldsymbol{W} \textrm{,}\]
	where $\odot$ computes the element-wise production.
	The mask operation bypasses the whole parameter firstly, then increases the weight of the selected part.
	We also add an auxiliary $L_0$ loss
	\[L_{\textrm{mask}} = \lambda_{\textrm{mask}} \sum_{l, \boldsymbol{W}} \lVert \boldsymbol{m}^l_{\boldsymbol{W}} \rVert _0 \textrm{,}\]
	which is a sparsity penalty to make the mask filter only the important components in a parameter.
	We follow~\citet{DBLP:conf/iclr/LouizosWK18} to optimize $L_0$ with their re-parametrization approach.
	It should be noted that the mask is only aware of and applied to the editing language because we aim to update all the languages simultaneously, making any assumption on the updating or retaining languages meaningless.
	
	Unfortunately, the element-wise masks for each language may contain as many parameters as the raw model, causing over-parameterization and a waste of computation.
	Say $h$ is the hidden size. If predicting the $O(h^2)$ updated parameters (or their gradients), the editor's parameters will inflate to unacceptable $O(h^4)$.
	Inspired by the capacity of the low-rank updating demonstrated in previous model editing work~\citep{DBLP:conf/emnlp/CaoAT21, DBLP:conf/iclr/MitchellLBFM22}, we factorize the full mask matrix into two low-rank matrics, then constructing the updated raw parameters with non-parameterized operations.
	
	The proposed \textit{language anisotropic} model editing can work with various model editing approaches, while the implementation is specific to the algorithm details.
	Taking a parameter matrix $\boldsymbol{W} \in \mathbb{R}^{n \times m}$ in an MLP layer for example. By the chain rule, its gradient on the loss $L$ is
	\[ \nabla_{\boldsymbol{W}} L = \boldsymbol{x}^\top \boldsymbol{\delta} \textrm{,}\]
	where $\boldsymbol{x} \in \mathbb{R}^n$ is the layer's input, and $\boldsymbol{\delta} \in \mathbb{R}^m$ refers to the gradient of the layer's output (i.e., the ``input'' in the backward pass).
	
	For hyper-network based approaches~\citep{DBLP:conf/emnlp/CaoAT21, DBLP:conf/iclr/MitchellLBFM22}, a network $g$ is built to conduct gradient transform. Hence, we insert the language masks $\boldsymbol{m}^l_{\boldsymbol{\cdot}}$ here as
	\[ \tilde{\boldsymbol{x}}, \tilde{\boldsymbol{\delta}} = g \left( \operatorname{mask} (\boldsymbol{x}, \boldsymbol{m}^l_{\boldsymbol{x}}), \operatorname{mask} (\boldsymbol{\delta}, \boldsymbol{m}^l_{\boldsymbol{\delta}}) \right) \textrm{.}\]
	
	For other approaches that do not manipulate gradients~\citep{DBLP:conf/iclr/SinitsinPPPB20}, the $g$ is an identical transformation, and the language masks do not affect the rest part of the editing algorithm.
	
	Finally we construct the full sized gradient using the rank-$1$ predictions
	\[ \tilde{\nabla}_{\boldsymbol{W}} L = \tilde{\boldsymbol{x}}^\top \tilde{\boldsymbol{\delta}} \textrm{.}\]
	
	The extra parameters and computation is in the order of $O(h \lvert \mathcal{L} \rvert)$. Since the size of the language set $\mathcal{L}$ is likely to be tens while the hidden size $h$ can easily reach the thousand level, the extra time-space cost is tiny compared to the original $O(h^2)$ order.
	To this end, we obtain an approach to make \textit{language anisotropic} model editing.
	
	\section{Experiments}

	\subsection{Evaluation}
	\label{sec:Experiments:Evaluation}
	
	To evaluate cross-lingual model editing approaches, we focus on \textit{cross-lingual transferability}, while continuing to keep our eyes on \textit{reliability} and \textit{locality}.
	Suppose that the languages we focus on make up $\mathcal{L}$, and the corpus is $\mathcal{D}_\mathcal{L}$.
	For $l \in \mathcal{L}$, each monolingual subset $\mathcal{D}_l$ of the corpus contains a number of tuples $(x_k, y_k)$, which means we desire the model to predict $y_k$ to the input $x_k$.
	The $y_k$ does not need to be different from the raw prediction $f(x_k; \boldsymbol{\theta})$.
	Taking the union of datasets in all the languages, we have the cross-lingual model editing dataset $\mathcal{D}_\mathcal{L} = \cup_{l \in \mathcal{L}} \mathcal{D}_l$.
	
	Inspired by \citet{DBLP:conf/emnlp/CaoAT21}, we propose three cross-lingual model editing metrics.
	Overall, we distinguish the languages where inputs are to be edited from where predictions are to be updated.
	Let $\mathcal{D}_{\textrm{edit}}$ be the set of (input, desired prediction) pairs fed to edit the model, which cause model predictions to inputs in $\mathcal{D}_{\textrm{update}}$ updated.
	In addition, $I(x) = \{x' \vert x'\ \textrm{is parallel to}\ x\}$ refers to parallel inputs of a specific input $x$ across languages of interest.
	
	To measure \textit{reliability} under \textit{cross-lingual transferability}, we use editing accuracy. We calculate the ratio of the predictions successfully updated
\[\operatorname{acc} = \mathbb{E}_{(x_e, y_e) \sim \mathcal{D}_{\textrm{edit}} \atop x_u \sim \mathcal{D}_{\textrm{update}} \cap I(x_e)} \left[ \mathds{1}_{f \left( x_u; \boldsymbol{\theta}_u \left( x_e, y_e \right) \right) = y_e} \right] \textrm{.}\]

To measure \textit{locality} under \textit{cross-lingual transferability}, we use editing consistency which reflects the retaining rate of predictions to unrelated inputs
\[\operatorname{con} = \mathbb{E}_{(x_e, y_e) \sim \mathcal{D}_{\textrm{edit}} \atop x_r \sim \mathcal{D}_{\textrm{update}} \setminus I(x_e)} \left[ \mathds{1}_{f \left( x_r; \boldsymbol{\theta}_u \left( x_e, y_e \right) \right) = f(x_r; \boldsymbol{\theta})} \right] \textrm{.}\]

The above two metrics are not necessarily consistent or even conflicting, similar to precision and recall in classification. Thus, we define the editing success rate as the harmonic mean
\[\operatorname{succ} = \frac{2 \times \operatorname{acc} \times \operatorname{con}}{\operatorname{acc} + \operatorname{con}} \textrm{.}\]

Since evaluating over the full set for each edit has a huge overhead of enumerating every two inputs, we follow existing work on model editing~\citep{DBLP:conf/emnlp/CaoAT21, DBLP:conf/iclr/MitchellLBFM22, DBLP:conf/icml/MitchellLBMF22} to estimate it with mini-batched expectation. Notably, in this work $\mathcal{D}_{\textrm{edit}}$ and $\mathcal{D}_{\textrm{update}}$ are finite datasets. Thus we enumerate each $(x_e, y_e) \in \mathcal{D}_{\textrm{edit}}$, and uniformly sample a subset in certain size of testing inputs $x_u$ from $I(x_e)$ or $x_r$ from complement set of $I(x_e)$ (for $\operatorname{acc}$ or $\operatorname{con}$, respectively) to make a pair in order to calculate the metrics.

To obtain an average metric over all the languages, we calculate the macro average over editing languages.
Specifically, to avoid enumerating all language pairs, we mix all the languages into $\mathcal{D}_{\textrm{update}} = \mathcal{D}_\mathcal{L}$, and use edit sets of single language from $\{ \mathcal{D}_l \} _{l \in \mathcal{L}}$ as $\mathcal{D}_{\textrm{edit}}$ one by one, then finally calculate the macro average.
Finally, the success rate is calculated using the averaged accuracy rate and consistency rate.

\begin{table*}[t]
	\centering
	\small
	\begin{tabular}{lccccccc}
		\toprule
		& & \multicolumn{3}{c}{\bf mLAMA} & \multicolumn{3}{c}{\bf XNLI}  \\
		\cmidrule(lr){3-5}\cmidrule(lr){6-8}
		\multicolumn{1}{c}{\bf Approach} & \bf Training Languages & \textbf{acc}\% & \textbf{con}\% & \textbf{succ}\% & \textbf{acc}\% & \textbf{con}\% & \textbf{succ}\%  \\
		\midrule
		Finetuning & n/a & 21.94 & 55.69 & 31.48 & 47.53 & 98.24 & 64.06  \\
		\hdashline\noalign{\vskip 0.5ex}
		Editable Training & en only & 51.13 & 17.33 & 25.88 & 71.02 & \bf 95.24 & 81.36  \\
		Editable Training & all & \bf 99.78 & \bf 24.45 & \bf 39.27 & \bf 89.45 & 93.04 & \bf 91.21  \\
		\hdashline\noalign{\vskip 0.5ex}
		KnowledgeEditor & en only & 37.18 & 50.19 & 42.72 & 69.96 & \bf 96.79 & 81.22  \\
		KnowledgeEditor & all & \bf 64.69 & \bf 53.00 & \bf 58.26 & \bf 86.20 & 95.08 & \bf 90.42  \\
		\hdashline\noalign{\vskip 0.5ex}
		MEND & en only & 24.76 & 61.09 & 35.24 & 84.90 & 94.87 & 89.61  \\
		MEND & all & \bf 99.58 & \bf 75.76 & \bf 86.05 & \bf 98.16 & \bf 97.75 & \bf 97.95  \\
		\bottomrule
	\end{tabular}
	\caption{Experiment results to compare monolingual and cross-lingual model editing approaches. During evaluation, the editing language is limited to en, while the updating and retaining languages contains all languages.}
	\label{tab:ex_m2x}
\end{table*}

\begin{figure*}
	\centering
	\includegraphics[scale=0.57]{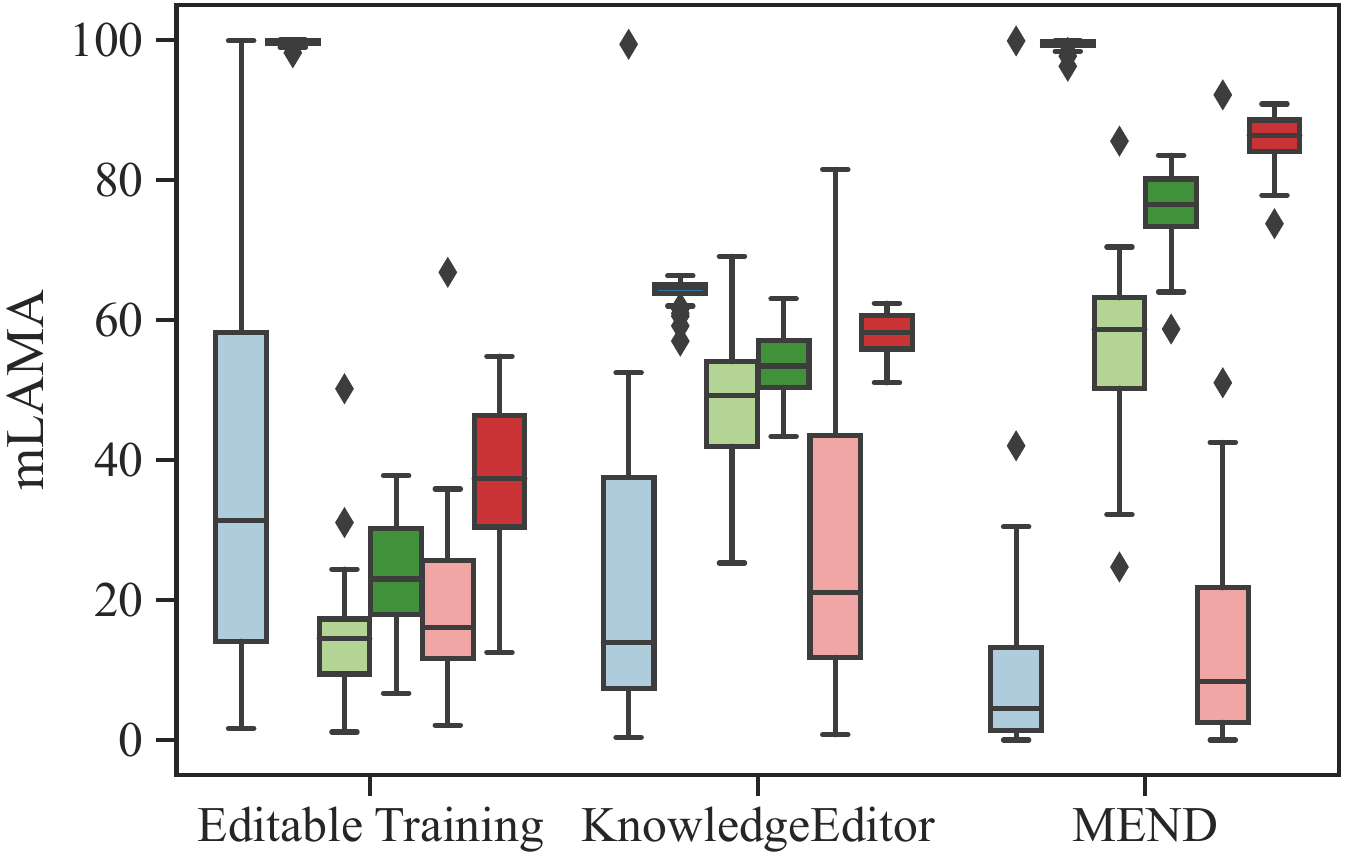}
	\hspace{0.6em}
	\includegraphics[scale=0.57]{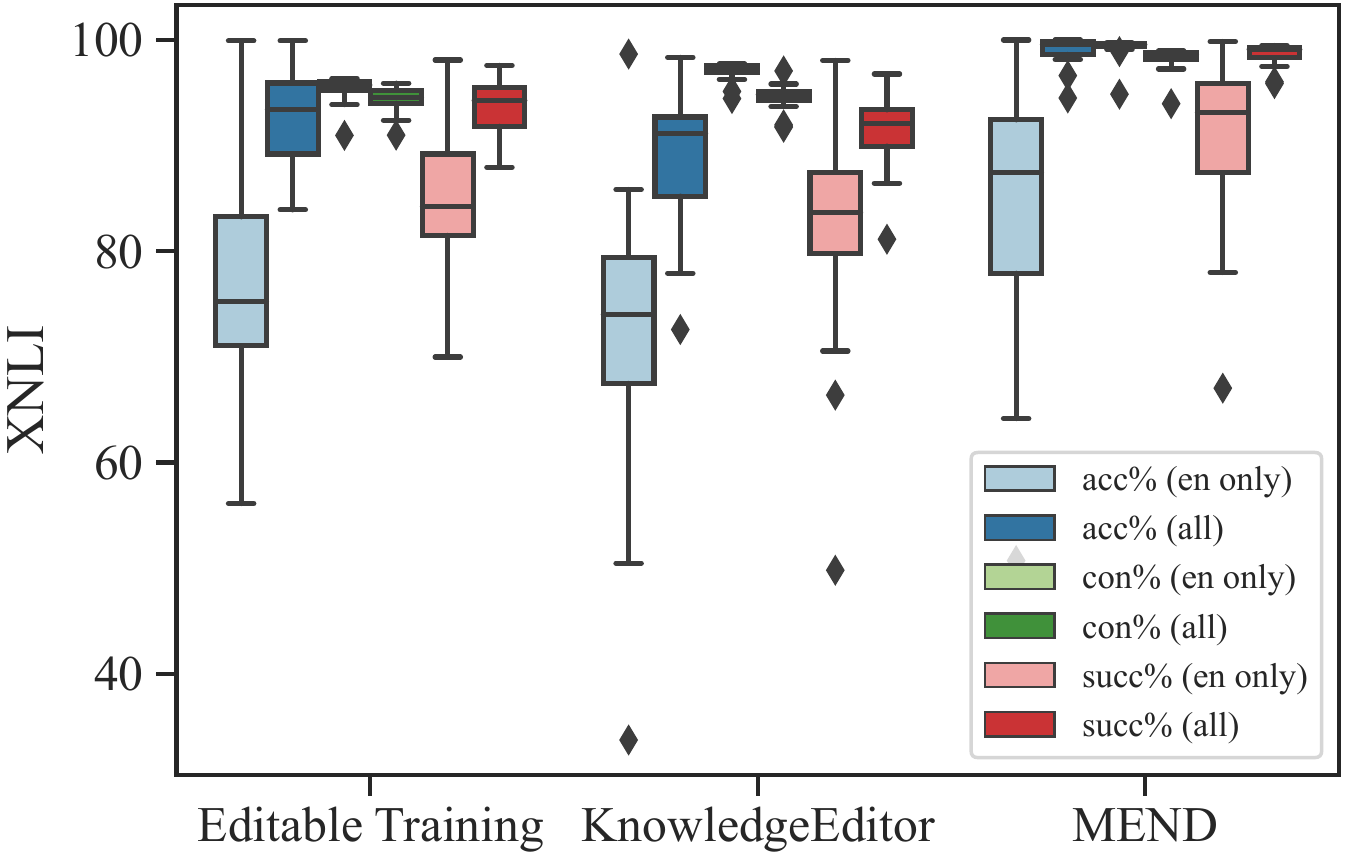}
	\caption{Editing performance varies across different languages. Training editors with parallel data improves overall editing performance, while decreasing the performance variance among languages.}
	\label{fig:ex_m2x_lg}
\end{figure*}

\subsection{Baselines}

\paragraph{Finetuning}
As the most common baseline of model editing, we use finetuning (degenerated editor).
With no editor to train, finetuning has no cross-lingual variant and makes no use of parallel corpus since no editor is to be trained.

\paragraph{Learned Editors}
Considering the proposed approaches are compatible with various learned editors, we use three monolingual editors as the basis: Editable Training~\citep{DBLP:conf/iclr/SinitsinPPPB20}, KnowledgeEditor~\citep{DBLP:conf/emnlp/CaoAT21}, and MEND~\citep{DBLP:conf/iclr/MitchellLBFM22}.
We compare the editing performance of each editor with and without our approaches.

\subsection{Datasets}
\label{sec:Experiments:Datasets}

Following the widely used setting, we construct synthetic editing datasets using existing data~\citep{DBLP:conf/iclr/SinitsinPPPB20, DBLP:conf/emnlp/CaoAT21, DBLP:conf/iclr/MitchellLBFM22, DBLP:conf/icml/MitchellLBMF22}.
We use the knowledge-intensive task mLAMA~\citep{DBLP:conf/eacl/KassnerDS21} for fact editing, which is natural because predictions involve only specific knowledge, which is prone to change.
Nevertheless, a usable dataset with a parallel corpus of another task, like classification, is lacking due to the difficulty in translating entities.
Therefore, to demonstrate the generic task-agnostic editing ability of cross-lingual model editing, we also use a semantics-focused dataset XNLI~\citep{DBLP:conf/emnlp/ConneauRLWBSS18} for error correction.

\paragraph{mLAMA}
is a multilingual dataset of knowledge probing task through (masked) language modeling, providing facts expressed as masked sentences in 53 languages.
Each fact is a triple $\langle \texttt{[X]}, \textrm{type}, \texttt{[Y]} \rangle$ including two entities, e.g., $\langle \textrm{Messi}, \textrm{play-for}, \textrm{Paris SG} \rangle$. To produce the textual expression from triples, mLAMA provides one template for each type (``play-for'') of fact like ``\texttt{[X]} plays for \texttt{[Y]}.''.

In the original setting of mLAMA, they fill the real \texttt{[X]} and replace \texttt{[Y]} with \texttt{[MASK]} tokens to probe the pre-trained language model.
In our model editing setting, to construct the editing input, we also keep the \texttt{[X]} but uniformly sample an entity within the same type as \texttt{[Y]}.
To measure the \textit{locality}, we replace the \texttt{[Y]} as \texttt{[MASK]} tokens in a row, where the number of \texttt{[MASK]} tokens is sampled from the length distribution of entity name in the corresponding language.
Note that translation of an entity may be invisible for the edited model or even nonexistent. Consequently, editing with entity names, which involves the entity linking problem, can be intractable in pure cross-lingual model editing.
Therefore, we always treat the entity in the edit input as the desired prediction.

\paragraph{XNLI}
is a parallel corpus of natural language inference in 15 languages, which can be modeled as a three-way sentence pair classification task, where we ask the model to predict the relation between a premise-hypothesis pair in $\{ \textrm{entailment}, \textrm{neutral}, \textrm{contradiction} \}$.

In the model editing scenario, we treat the premise-hypothesis pair as a whole input sentence to classify.
Unfortunately, since the raw model has already been finetuned using the training and dev set, a dedicated training setting for error correction cannot be built.
Thus, we train the editor to edit arbitrarily, which implies the error correction ability.
During training, we sample edit input over the training set and give a uniformly random label as the desired prediction.
To evaluate an editor on \textit{reliability}, we use data in the test set that the raw model gives wrong predictions and use corresponding gold labels as the desired predictions.
As for \textit{locality}, we continue to sample inputs to be retained over the whole test set.

\begin{table*}[t]
	\centering
	\small
	\begin{tabular}{lcccccc}
		\toprule
		& \multicolumn{3}{c}{\bf mLAMA} & \multicolumn{3}{c}{\bf XNLI}  \\
		\cmidrule(lr){2-4}\cmidrule(lr){5-7}
		\multicolumn{1}{c}{\bf Approach} & \textbf{acc}\% & \textbf{con}\% & \textbf{succ}\% & \textbf{acc}\% & \textbf{con}\% & \textbf{succ}\%  \\
		\midrule
		Fintuning & 10.14 & 48.68 & 16.79 & 56.48 & 98.54 & 71.81  \\
		\hdashline\noalign{\vskip 0.5ex}
		Editable Training & 97.39 & 21.90 & 35.75 & 90.02 & 93.58 & 91.76  \\
		\quad w/ \textit{Language Anisotropic} Model Editing & \bf 97.87 & \bf 24.41 & \bf 39.08 & \bf 91.79 & \bf 93.68 & \bf 92.72  \\
		\hdashline\noalign{\vskip 0.5ex}
		KnowledgeEditor & 47.30 & 49.32 & 48.29 & 83.88 & \bf 95.79 & 89.44  \\
		\quad w/ \textit{Language Anisotropic} Model Editing & \bf 55.91 & \bf 51.00 & \bf 53.34 & \bf 86.88 & 95.45 & \bf 90.96  \\
		\hdashline\noalign{\vskip 0.5ex}
		MEND & 94.83 & 67.59 & 78.92 & 98.16 & 97.44 & 97.80  \\
		\quad w/ \textit{Language Anisotropic} Model Editing & \bf 96.12 & \bf 69.20 & \bf 80.47 & \bf 98.42 & \bf 98.02 & \bf 98.22  \\
		\bottomrule
	\end{tabular}
	\caption{Experiment results show that all three editors benifit from \textit{language anisotropic} model editing on both datasets. All of the approaches are trained and evaluated in all languages.}
	\label{tab:ex_lae}
\end{table*}

\begin{figure*}
	\centering
	\includegraphics[scale=0.57]{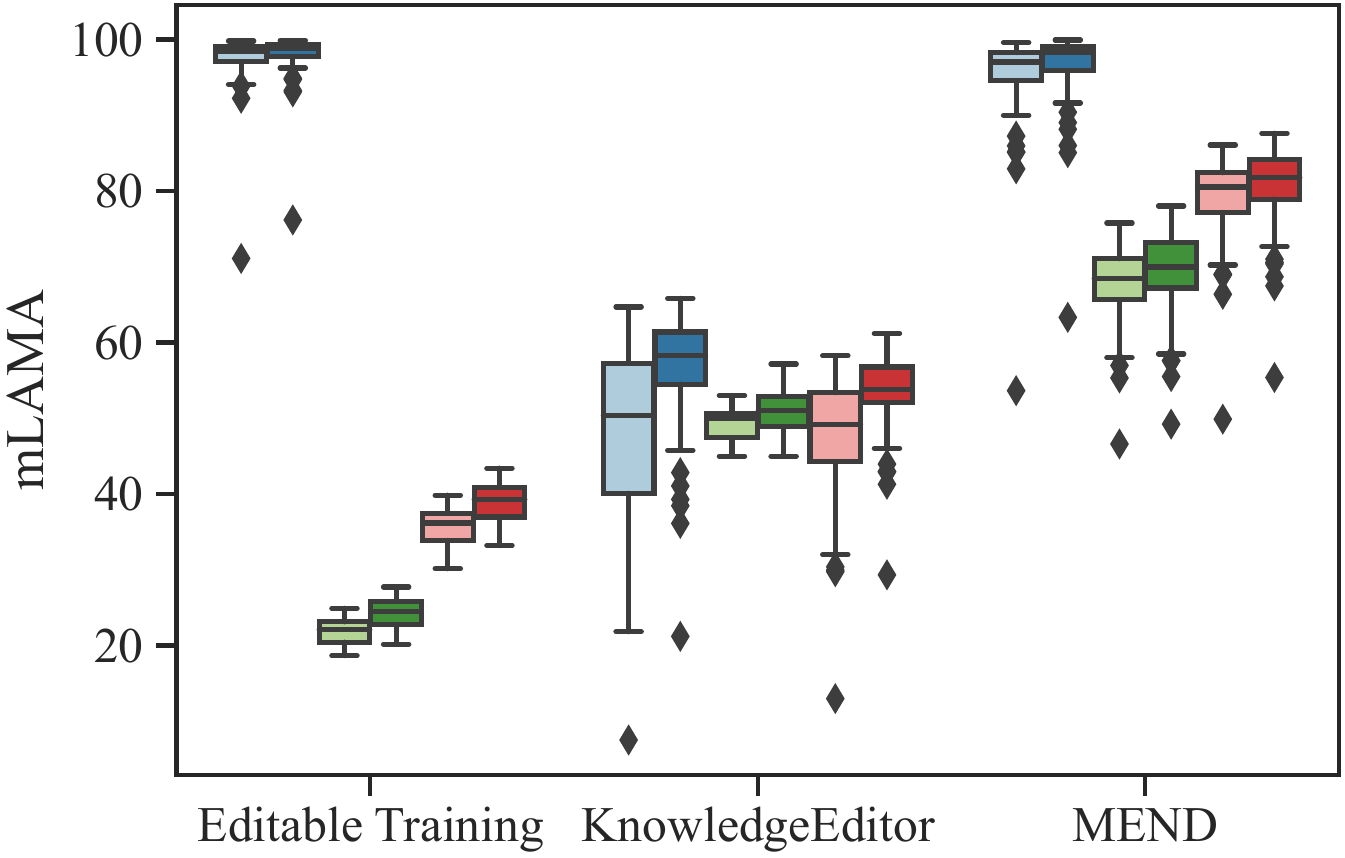}
	\hspace{0.6em}
	\includegraphics[scale=0.57]{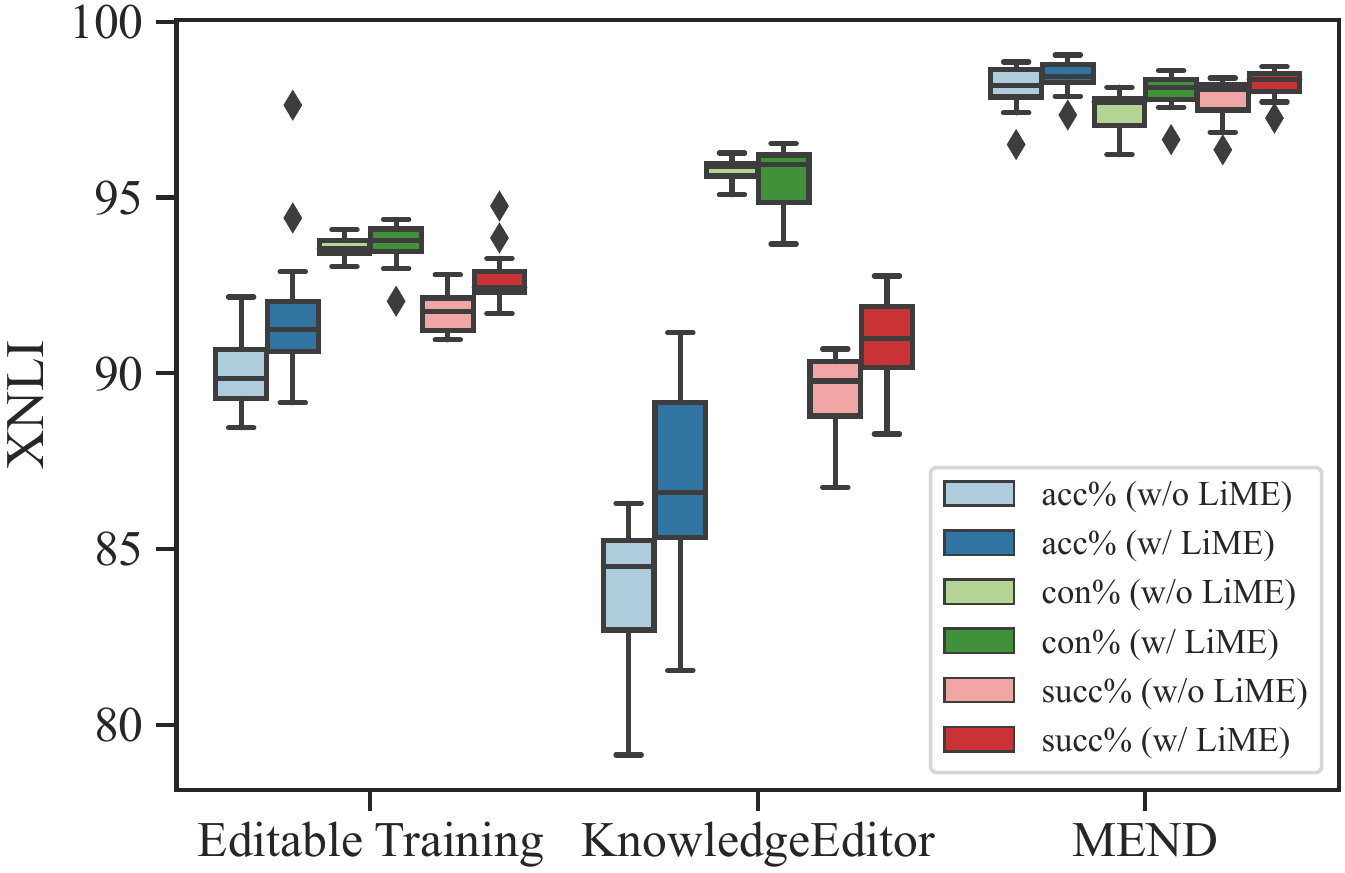}
	\caption{Distribution of editing performance across languages. \textit{Language Anisotropic} Model Editing (LiME for short) provides overall performance improvement and closes the performance gap across languages.}
	\label{fig:ex_lae_lg}
\end{figure*}

\subsection{Cross-Lingual Model Editing}
\label{sec:Experiments:Cross-Lingual Model Editing}

In this part, we demonstrate that the cross-lingual scenario exceeds the capability of the monolingual model editing approach.
Specifically, we compare the editing performance of the monolingual approaches and the proposed cross-lingual variants.
Recall that we use $\mathcal{L}$ to represent the full language set, i.e., the 15 languages for XNLI and the 53 for mLAMA.
In the case of XNLI, the data is inherently parallel, while in mLAMA, each language, excluding English, relies on a translated subset of English.
Given this scenario, we train the editors using the English subset to ensure uniform exposure to knowledge during training, thereby mitigating potential issues arising from training set disparities.
More specifically, we select en as the editing language and expect the approaches to update predictions across all the languages. Hence, we have $l_e = \textrm{en}$, $l_u, l_r \in \mathcal{L}$ during evaluation.
Table~\ref{tab:ex_m2x} shows the averaged results of en $\rightarrow$ all the languages, while Figure~\ref{fig:ex_m2x_lg} illustrates the distribution of results across languages.

Finetuning suffers from severe cross-lingual underfitting according to its low editing accuracy, causing a low overall success rate.

The monolingual editors work much better than finetuning. Although never seen other languages, the editors demonstrate partial \textit{cross-lingual transferability}. Moreover, the editor acquires the ability to perform updates of \textit{locality}, almost reaching all the highest editing consistency on XNLI.

However, only editors with the proposed cross-lingual editing training framework truly generalizes the desired prediction to inputs in other languages. On XNLI, editors trained cross-lingually on all languages improves the editing accuracy by a large margin, with much less loss of editing consistency, resulting in a large growth in editing success rate. On mLAMA, where the model faces a much larger output space, editors trained on all languages reveals its high consistency and improves all three metrics significantly. Moreover, Figure~\ref{fig:ex_m2x_lg} shows that the performance gap across languages is closer under the cross-lingual training framework.

\subsection{Language Anisotropic Model Editing}
\label{sec:Experiments:Language Anisotropic Model Editing}

After confirming the effectiveness of the cross-lingual model editing, we conduct experiments to study how the proposed \textit{language anisotropic} model editing improves performance.
Here we always train and evaluate approaches in all languages ($l_e, l_u, l_r \in \mathcal{L}$).
Table~\ref{tab:ex_lae} shows the averaged all $\rightarrow$ all results, with the per-language distribution plotted in Figure~\ref{fig:ex_lae_lg}.
Editors using parallel training data in Table~\ref{tab:ex_m2x} are the same as the editors without \textit{language anisotropic} model editing in Table~\ref{tab:ex_lae}.
The difference is that we no longer limit the editing language, thus the editing task becomes harder, making the results in Table~\ref{tab:ex_lae} lower.

Finetuning still falls into underfitting across languages, performing similar to the situation of single editing language.

With \textit{language anisotropic} model editing, the performance of editors reach a new high in both datasets.
Note that on XNLI, the small growth ($97.80\% \rightarrow 98.22\%$) corresponds to the large error reduction ($2.20\% \rightarrow 1.78\%$, $19\%$ relatively).

Though trained with parallel data, a performance gap still exists between some languages and the others.
\textit{Language anisotropic} model editing helps the editors close the performance gap and increases the overall editing success rates.

\begin{figure}[t]
	\centering
	\includegraphics[width=\columnwidth]{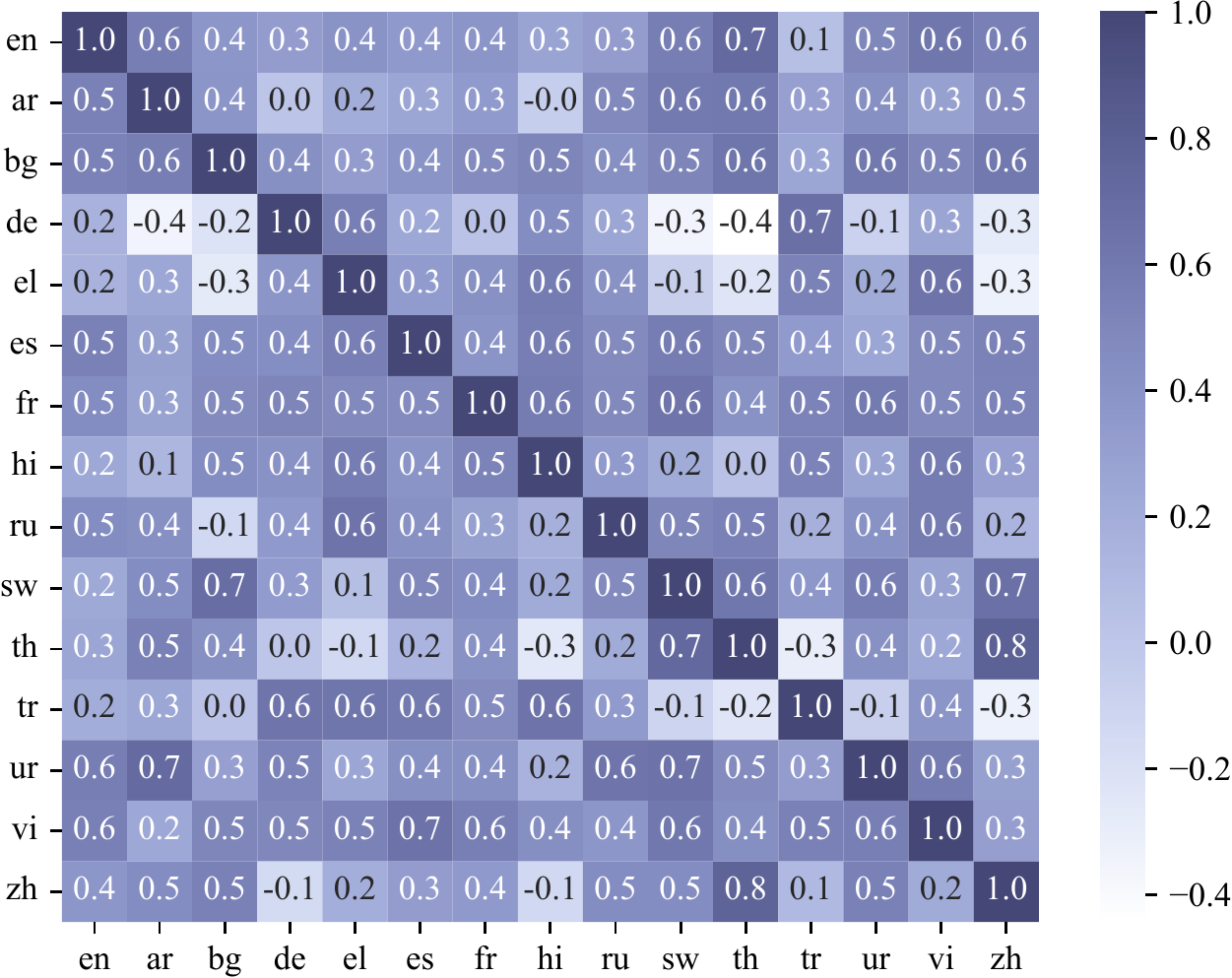}
	\caption{Cosine similarities of learned parameters of language-specific masks. In each row, we inspect the top-$1\%$ preferred dimensions of a certain language $l$ by value, on which dimensions we calculate the cosine similarities between $l$ and every languages.}
	\label{fig:heatmap}
\end{figure}

To illustrate the function of the language-specific masks, we conduct analyses using one of the final MEND based checkpoints on XNLI.
We observe that the parameters of the masks are very close in most dimensions across all languages. However, masks for different languages show preferences in small but different dimension subsets.
Therefore, we plot the cosine similarities of learned parameters in the masks as a heatmap in Figure~\ref{fig:heatmap}, where we limit the size of the preferred subset to $1\%$ of the full hidden size.
The heatmap of cosine similarities demonstrates that \textit{language anisotropic} model editing captures the \textit{language anisotropy} feature of the multilingual pre-trained language model.
Through adaptively re-weighting gradients of a small subset of parameters for each language, \textit{language anisotropic} model editing improves the performance of cross-lingual model editing.

\section{Related Work}

\paragraph{Model Editing}
\citet{DBLP:conf/iclr/SinitsinPPPB20} initially presents the model editing problem and proposes a MAML-like method, called Editable Training.
Our cross-lingual model editing problem definition and metrics mostly extend their work.
The proposed \textit{language anisotropic} model editing approach can be applied to Editable Training by using the rank-$1$ masks to construct a full gradient/parameter mask.

A series of work models editing as a learning-to-update problem and develops the hyper-network based approaches, such as KnowledgeEditor~\citep{DBLP:conf/emnlp/CaoAT21}, MEND~\citep{DBLP:conf/iclr/MitchellLBFM22}, and SLAG~\citep{DBLP:journals/corr/abs-2111-13654}.
They build the editor to constrain gradients during finetuning. We gain a lot of inspiration from their work when designing our methods.

A category of approaches regard the language model as a knowledge base, and utilize a wider range of editing fomulars~\cite{DBLP:conf/nips/SanturkarTEBTM21, DBLP:journals/corr/abs-2202-05262, DBLP:journals/corr/abs-2210-07229, DBLP:conf/emnlp/GevaSBL21, dai-etal-2022-knowledge, DBLP:conf/icml/MitchellLBMF22}.
We can obtain the cross-lingual variants using the parallel corpus, while whether the \textit{language anisotropic} model editing works depends on the algorithm details.

\paragraph{Cross-Lingual Transferability}
In recent work, multilingual pre-trained language models show their \textit{cross-lingual transferability}~\citep{DBLP:conf/naacl/DevlinCLT19, DBLP:conf/acl/ConneauKGCWGGOZ20, DBLP:conf/naacl/XueCRKASBR21, DBLP:conf/naacl/ChiDWYSWSMHZ21}, where multiple languages included in the training corpus benifit each other.
Opposite to the positive cross-lingual transfer,~\citet{DBLP:conf/emnlp/WangLT20} study the negative interference phenomenon.
They show the existence of language-specific parameters, which is also a theoretical basis of our work.
Based on this priori, their work and our proposed \textit{language anisotropic} model editing have similar underlying ideas: identifying the language-specific parameters and using them to improve the \textit{cross-lingual transferability}.

Though, our work differs from theirs in method and task.
They leverage language-specific pruning to identify the preferred parameter subset of different languages. Then they propose an iterative second-order meta-optimizing algorithm to improve pre-training.
Our approach does not perform prune, where the masks play the role of reweighting coefficients. Our approach also makes no change in the training algorithm, maintaining maximum compatibility with various model editing approaches.

\section{Conclusion}
In this work, we define the task and metrics of cross-lingual model editing.
After summarizing the training formula of various monolingual model editing approaches, we naturally extend the formula to a cross-lingual variant using the parallel corpus.
Further, we propose \textit{language anisotropic} model editing to improve cross-lingual model editing.
We conduct experiments to verify that the cross-lingual model editing problem is necessary and find that the proposed approaches are effective.

\section*{Limitations}
Our work depends mainly on parallel data. Although tasks focusing on language abilities can leverage machine translation to obtain parallel data~\citep{DBLP:journals/corr/abs-2003-11080}, it is much harder for tasks about knowledge and facts.
Using parallel data to train cross-lingual model editors is like doing full supervision, while we need to leverage weakly labeled data to mitigate data scarcity.

On the other hand, whether monolingual or cross-lingual, model editing still struggles with the continual learning problem.
In the real world, knowledge constantly emerges and fades, disabling the stop of learning.
However, most studies, including our work, focus on a single or a batch of inputs.
Thus, an effective solution of continuously updating a series of inputs is necessary before model editing becomes a practical technic.

Note that our work focuses on the editor's generalized cross-lingual editing ability. We expect the editor to perform the editing honestly. This target potentially offers the possibility to modify model behavior maliciously. Though editing may not soon become a practical technic, the potential risk does exist.

\section*{Acknowledgement}
This work was supported by the National Key R\&D Program of China via grant 2020AAA0106501 and the National Natural Science Foundation of China (NSFC) via grant 62236004 and 61976072.

\bibliography{custom}
\bibliographystyle{acl_natbib}

\appendix

\section{Datasets Preprocessing}

\subsection{mLAMA}

Along with the raw English data from LAMA~\citep{DBLP:conf/emnlp/PetroniRRLBWM19}, mLAMA provides translations of the facts in the other 52 languages if possible.
mLAMA is organized into two-level. The first level is relations, and the second is facts.
Facts in the same relation share the same template and can be identified by the $\langle \texttt{[X]}, \texttt{[Y]} \rangle$.
Thus, we split at the fact level for consistency across different data splits.
Specifically, we split the whole dataset to train/dev/test with a ratio of 8:1:1, resulting in 628,612/78,555/78,600 facts in the train/dev/test set.

In our setting, the template with filled \texttt{[X]} is used as input, and then we test if the edited model predicts \texttt{[Y]} (\S\ref{sec:Experiments:Datasets}).
Thus, to avoid leakage and keep large output space, we ensure that an \texttt{[X]} can only appear in one split while not limiting the label \texttt{[Y]}.
Take an example of relation P19 (\texttt{[X]} was born in \texttt{[Y]} .) where \texttt{[X]} is a person's name and \texttt{[Y]} is a location.
In the training set, if an input is ``\texttt{Allan Peiper} was born in \texttt{[MASK]} .'', there cannot exist an input in the dev/test set with $\texttt{[X]} = \texttt{Allan Peiper}$.
On the contrary, a \texttt{[Y]} (like ``\texttt{Alexandra}'') can be used as the desired prediction during both training and testing, because we use it as a label.
We first exclude samples in the test set having overlap \texttt{[X]} with the training set, then exclude samples in the dev set overlapping with the train/test.
Finally, we obtain 628,612/23,718/53,993 facts in the train/dev/test set after preprocessing summing up all languages.

\subsection{XNLI}

XNLI dataset contains completely parallel data in fifteen languages.
We use Huggingface Datasets to access to XNLI dataset, and follow the official split with 392,702/2,490/5,010 samples in train/dev/test set in each language.

\section{Experiment Details}

We conduct all experiments three times and use the mean of editing success rates as the final performance metric, including main experiments and hyperparameter tuning.

\subsection{Model and Implementation}

We use \texttt{bert-base-multilingual-cased} from Hugging Face Transformers~\citep{DBLP:conf/emnlp/WolfDSCDMCRLFDS20} as the basic model.
As the basic model editors, we use MAML-like Editable Training~\citep{DBLP:conf/iclr/SinitsinPPPB20}, and hyper-network based KnowledgeEditor~\citep{DBLP:conf/emnlp/CaoAT21} and MEND~\citep{DBLP:conf/iclr/MitchellLBFM22}.
For XNLI, the pre-trained model finetuned on the en training set is used as the raw model to edit in all the following experiments.
For mLAMA, we use the pre-trained language model directly.

We work on the official MEND codebase, together with HuggingFace Transformers and Datasets.
During preliminary experiments, we find that MEND in their implementation suffers from low computation efficiency and sub-optimization due to the token-level BatchNorm~\citep{DBLP:conf/icml/IoffeS15} variant used by the editor. Thus, we replace the token-level BatchNorm in MEND editor with LayerNorm~\citep{DBLP:journals/corr/BaKH16} in our implementation.

\subsection{Hyperparameters}

\paragraph{Finetuning}
We use Adam~\citep{DBLP:journals/corr/KingmaB14} optimizer with learning rate of $5 \times 10^{-6}$.
For each input to be edited, the maximum step is set to $100$.

\paragraph{Learned Editors}
We follow~\citet{DBLP:conf/iclr/MitchellLBFM22} in most of the hyperparameter settings of the three monolingual editing approaches we use.
For Editable Training and KnowledgeEditor, we set the learning rate to $5 \times 10^{-5}$.
For MEND, the editor is initialized to an identical mapping and trained by Adam optimizer with the learning rate of $1 \times 10^{-6}$.
For the inner gradient decent updating, the learning rate is set to $1 \times 10^{-4}$.
For the coefficients of losses, we set $\lambda_{\textrm{rel}} = 0.1$ and $\lambda_{\textrm{loc}} = 1.0$.

Since \texttt{bert-base-multilingual-cased} is used as the raw model, we follow the setting of \texttt{bert-base} of the original MEND, i.e., editing MLPs in the last three layers of the encoder, leaving the other parameters frozen.

\paragraph{Language Anisotropic Model Editing}
The \textit{language anisotropic} varients inherit hyperparameters for architectures and training from their corresponding base editors.

The newly introduced training hyperparameters include the learning rate of masks and $\lambda_{\textrm{mask}}$.
We tune the learning rate of masks in $\{ 1 \times 10^{-4}, 1 \times 10^{-3}, 1 \times 10^{-2} \}$, and $\lambda_{\textrm{mask}}$ in $\{ 0.01, 0.1, 1 \}$.
We pick the best values and apply them to all main experiments, i.e., learning rate of masks of $1 \times 10^{-3}$, and $\lambda_{\textrm{mask}}$ of $1.0$.

\subsection{Training Details}

We utilize the early-stopping strategy along with up to $500,000$ training steps.
When training on the full datasets, we evaluate the model every $100,000$ steps and finalize training when the editing success rate is not improved over $200,000$ steps.
When training on the English only subset, the validation interval is set to $20,000$ and the early stop patience is $40,000$ steps.

All experiments fit in one NVIDIA RTX 2080Ti GPU, where a single run takes one to three days.

\section{Additional Results}

The large versions with raw data points of Figure~\ref{fig:ex_m2x_lg} and Figure~\ref{fig:ex_lae_lg} are as follows.

\begin{figure*}
	\centering
	\includegraphics[scale=1.15]{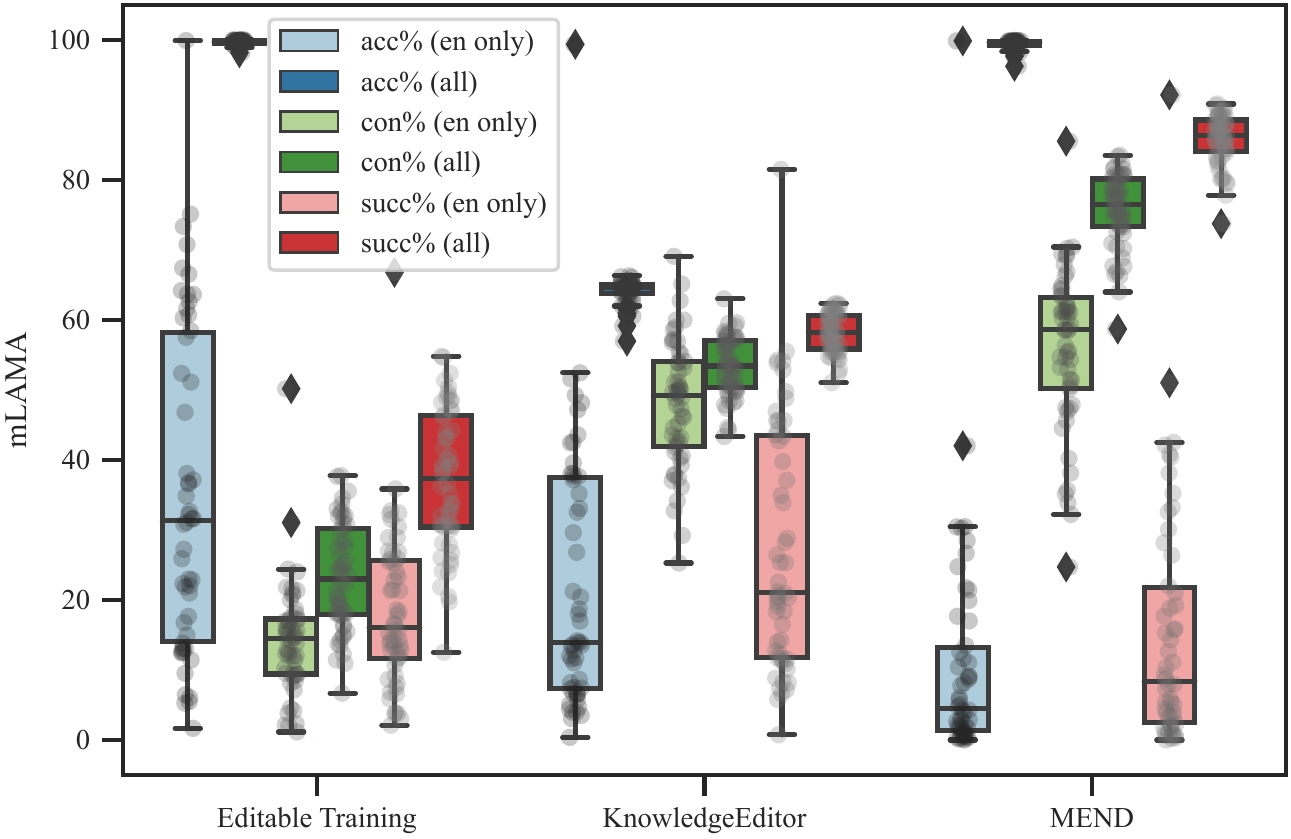}
	\includegraphics[scale=1.15]{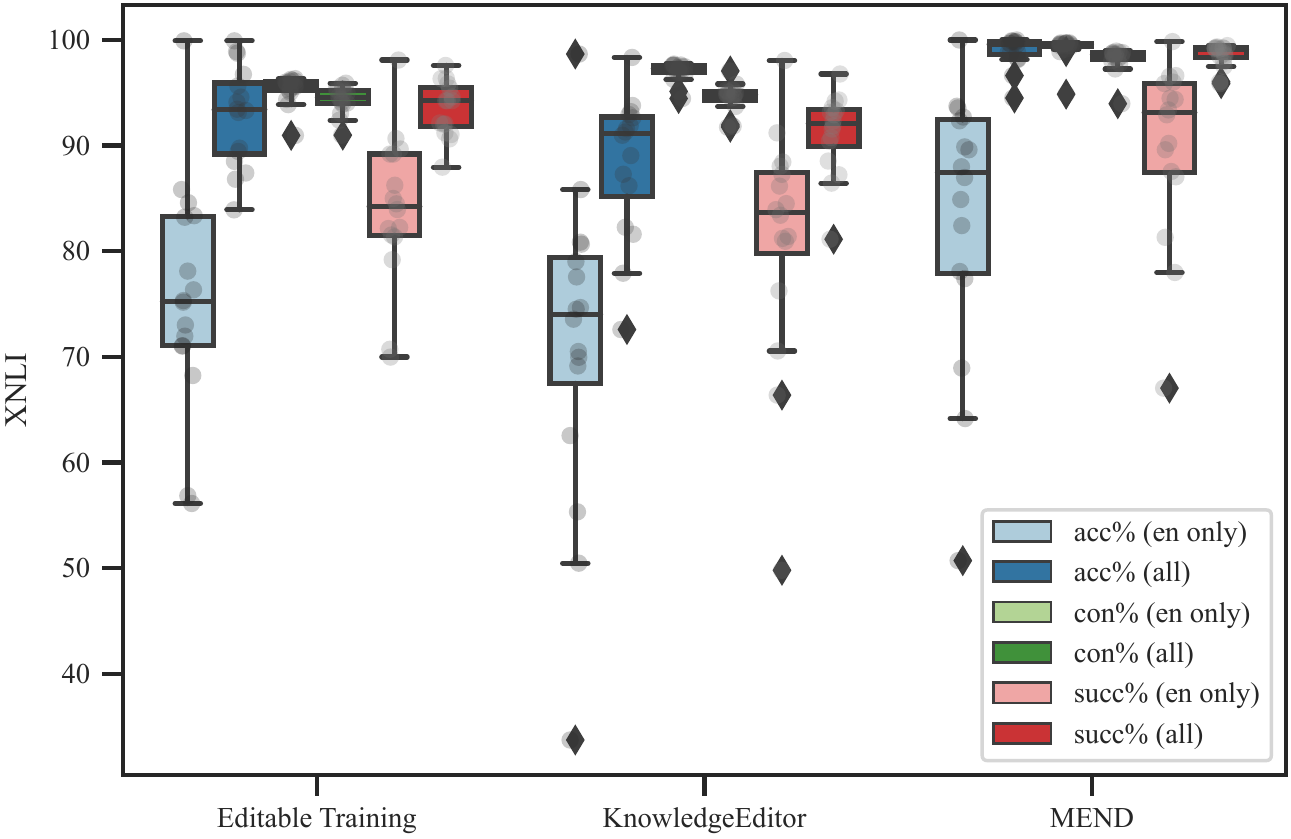}
	\caption{A large version of Figure~\ref{fig:ex_m2x_lg}, where each grey point corresponds to  an en $\rightarrow$ $l$ result averaged over three runs.}
	\label{fig:ex_m2x_lg_large}
\end{figure*}

\begin{figure*}
	\centering
	\includegraphics[scale=1.15]{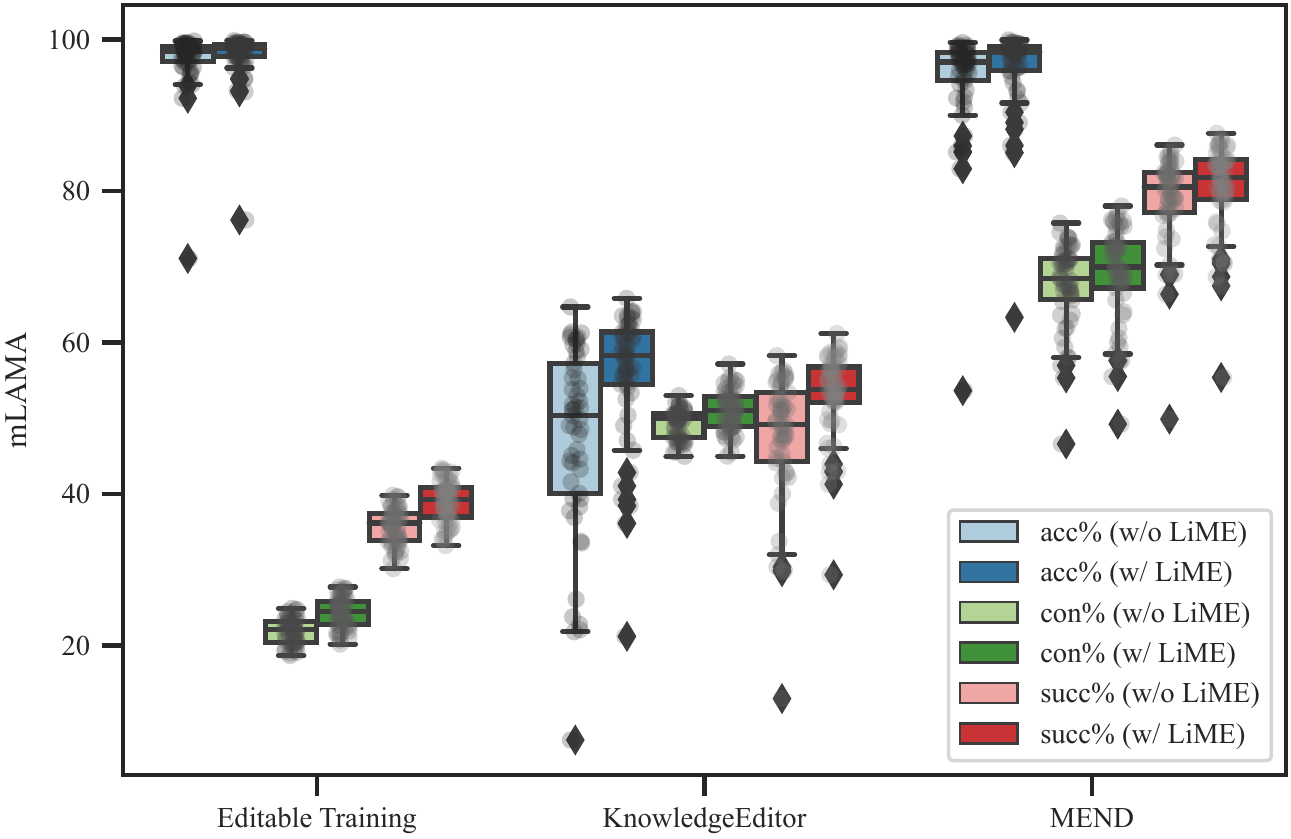}
	\includegraphics[scale=1.15]{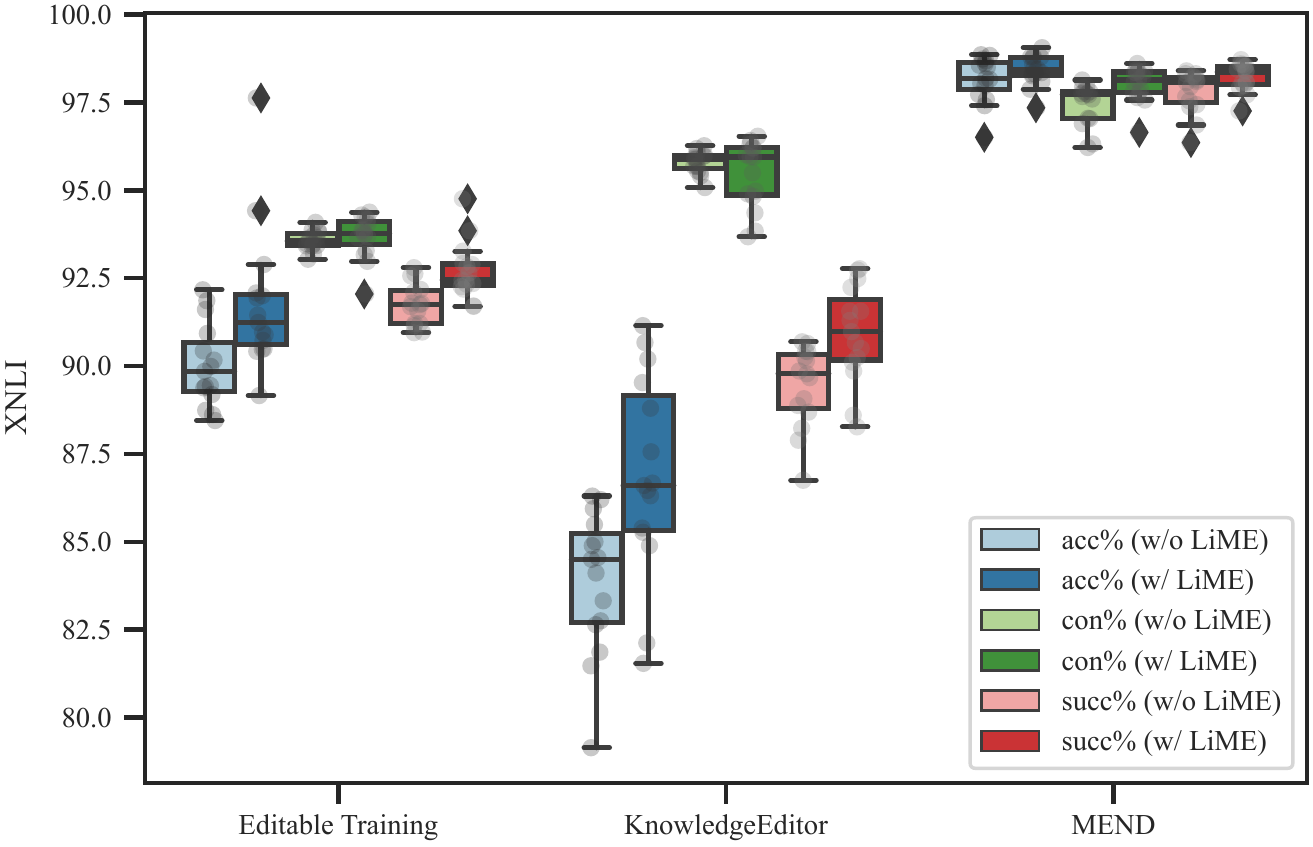}
	\caption{A large version of Figure~\ref{fig:ex_lae_lg}, where each grey point corresponds to  an $l$ $\rightarrow$ all result averaged over three runs. LiME for \textit{Language Anisotropic} Model Editing.}
	\label{fig:ex_lae_lg_large}
\end{figure*}

\end{document}